\pdfoutput=1
\pdftrailerid{}
\documentclass{article}
\usepackage{vgym,times}
\usepackage[T1]{fontenc}
\usepackage[utf8]{inputenc}
\usepackage{amsmath,amssymb}
\usepackage{graphicx,booktabs}
\usepackage{placeins}
\usepackage{wrapfig}
\usepackage{microtype}


\usepackage{multirow}
\usepackage[table]{xcolor}
\usepackage[skins,breakable]{tcolorbox}
\definecolor{prompttitle}{HTML}{D8EAF7}
\definecolor{promptbody}{HTML}{F4F9FD}
\definecolor{promptborder}{HTML}{8AB6D6}
\definecolor{promptink}{HTML}{1F4E79}
\newtcolorbox{promptbox}[1]{%
  enhanced,breakable,colback=promptbody,colframe=promptborder,
  colbacktitle=prompttitle,coltitle=promptink,
  fonttitle=\normalsize\bfseries,fontupper=\small,
  title={#1},title after break={#1\ (continued)},
  boxrule=0.6pt,arc=2mm,left=9pt,right=9pt,top=7pt,bottom=7pt,
  toptitle=5pt,bottomtitle=5pt,before skip=8pt,after skip=10pt,
  before upper={\raggedright\setlength{\parindent}{0pt}\setlength{\parskip}{4pt}}}
\newcommand{\promptfield}[1]{\par\smallskip\textbf{#1}\par\nobreak}
\newcommand{\promptvar}[1]{\texttt{\{\detokenize{#1}\}}}
\usepackage{algorithm,algorithmic}
\usepackage{hyperref,url}
\hypersetup{
  hidelinks,
  pdftitle={V-Gym: Enhancing Agentic Visual Reasoning via Skill-Data Co-Evolution},
  pdfauthor={Bei Yan, Yuecong Min, Jie Zhang, Junqi Yang, Shiguang Shan, Xilin Chen},
  pdfsubject={Agentic visual reasoning and skill-data co-evolution},
  pdfkeywords={multimodal agents, visual reasoning, skill evolution, data evolution}
}

\title{V-Gym: Enhancing Agentic Visual Reasoning\\via Skill-Data Co-Evolution}
\author{%
Bei Yan\textsuperscript{1,2}, Yuecong Min\textsuperscript{1,2}, Jie Zhang\textsuperscript{1,2}, Junqi Yang\textsuperscript{1,2}, Shiguang Shan\textsuperscript{1,2}, and Xilin Chen\textsuperscript{1,2}\\[3pt]
\textsuperscript{1} State Key Laboratory of AI Safety, Institute of Computing Technology,\\
Chinese Academy of Sciences, Beijing, China\\
\textsuperscript{2} University of Chinese Academy of Sciences, Beijing, China}
\date{}

\begin{document}
\maketitle
\suppressfloats[t]

\begin{abstract}
Advances in multimodal understanding, reasoning, and tool use enable agents to tackle increasingly complex visual reasoning tasks. By distilling past execution experience into reusable skills, agents can transfer lessons from both successes and failures into future reasoning, reducing repeated errors and improving capabilities. However, limited experience may produce unreliable, poorly generalizable skills, while static datasets may lack the targeted and diverse practice needed for refinement. To address this gap, we introduce V-Gym, an autonomous framework that iteratively co-evolves procedural skills and multimodal practice data from execution trajectories. \textit{During skill evolution}, V-Gym analyzes trajectories to distill and refine hierarchical skills, updating procedural guidance and applicability conditions while retaining an update only if it improves validation performance. \textit{During data evolution}, V-Gym selects generation seeds by balancing data utility and exploration, then translates trajectory-identified bottlenecks into diverse, targeted practice data that expand the data bank after quality checks. The resulting practice outcomes feed back into subsequent skill updates, closing the loop for continual skill refinement. Experiments across diverse multimodal reasoning benchmarks show substantial improvements over baselines with multiple backbone models. Its evolved skills generalize across domains and models, while evolved data support more effective skill refinement, enabling autonomous diagnosis, targeted practice, and continual self-improvement.

\end{abstract}

\section{Introduction}
\label{sec:introduction}
\begin{figure}[t]
    \centering
    \includegraphics[width=\linewidth]{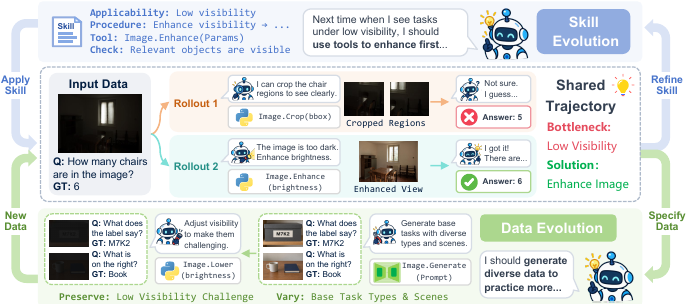}
    \caption{\textbf{Shared trajectories for skill-data co-evolution.}
    They guide skill refinement (top) and targeted data generation (bottom), updating both banks for later practice.}
    \label{fig:trajectory}
\end{figure}

Advances in multimodal understanding, reasoning, and tool use increasingly enable agents to solve complex tasks by coordinating visual observations, evidence integration, and multistep actions~\citep{surís2023vipergptvisualinferencepython,hu2024visualsketchpadsketchingvisual}. Beyond solving individual tasks, these agents also need to learn from experience to reduce recurring errors and adapt to unfamiliar conditions. A central challenge for such self-improvement is how to transform execution experience into reusable knowledge that benefits subsequent tasks. Prior work has explored distilling execution trajectories into skills, i.e., procedural knowledge that specifies action sequences, applicability conditions, and verification rules~\citep{wang2023voyageropenendedembodiedagent,wang2024agentworkflowmemory,jiang2026xskillcontinuallearningexperience}. By retrieving and applying these skills, agents can leverage accumulated experience to guide future decisions.

Execution-derived skills are hypotheses about how to solve a task, rather than verified rules~\citep{yang2026skilloptexecutivestrategyselfevolving}. A successful rollout shows that a procedure worked in one context without revealing which visual cues or tool states were essential, while a failed rollout may conflate a flawed procedure with missing evidence or incidental execution noise. Reliable skill evolution therefore requires targeted, varied practice to consolidate procedures, test applicability, and revise them using new evidence. The supporting practice data\footnote{In this paper, \textit{practice data} denote original or quality-checked synthetic data instances selected for agent practice and used to collect trajectories and feedback for skill and data evolution.} should, in turn, adapt to evolving skill-refinement needs, which fixed datasets may not adequately address~\citep{huang2026onpolicydataevolutionvisualnative}. This challenge is particularly pronounced in multimodal tasks, where meaningful variation must preserve consistency among visual content, questions, and answers while retaining the evidence needed to exercise the intended reasoning and tool-use operations~\citep{luo2024mmevolempoweringmultimodallarge,zeng2026makesgoodagenticdata}. Multimodal generation can construct such practice data autonomously and at scale~\citep{alam2026sphinxsyntheticenvironmentvisual,huang2026onpolicydataevolutionvisualnative}, but generation alone does not identify which task bottlenecks future practice should target or which factors can be varied without weakening the intended challenge. This motivates a coupled self-improvement process in which execution evidence jointly determines how skills are refined and which practice data are selected or generated next.

Our key insight is that \textit{execution trajectories provide a shared foundation for both skill and data evolution}. As illustrated in Figure~\ref{fig:trajectory}, execution trajectories reveal reusable procedures and failure patterns that guide skill refinement, while identifying the core task challenges that new practice data should preserve and the aspects that can be varied to introduce diversity. Updated skills and generated data support future practice. We propose V-Gym, an autonomous practice framework for improving agentic visual reasoning through skill-data co-evolution.

V-Gym retrieves relevant skills and performs tool-assisted reasoning on practice data, collecting execution trajectories and evaluation feedback. \textit{During skill evolution}, it groups related trajectories and analyzes each group to propose skill patches that revise procedural guidance and applicability conditions. Patches targeting the same skill object are consolidated into a single candidate skill update, which is accepted only if it improves task-solving performance on a validation set. \textit{During data evolution}, it attributes skill-update gains to contributing data instances and balances this utility with exploration to select generation seeds. Analysis of the selected seeds' trajectories identifies the core task challenges to preserve and the aspects that can be varied, guiding tool-assisted, targeted practice data generation. Generated instances that pass model self-checks expand the data bank for subsequent practice, providing further opportunities for skill refinement.

We evaluate V-Gym with advanced large vision-language model (LVLM) backbones on diverse benchmarks for complex multimodal reasoning. Extensive experiments show that V-Gym consistently outperforms baselines across the evaluated backbones and multimodal scenarios. The evolved skills also generalize to unseen domains and transfer effectively to less capable backbone models, highlighting the transferability of procedural knowledge acquired through autonomous practice. Moreover, continued practice with evolving data leads to more effective skill refinement than practice with a fixed dataset. These results support the potential of jointly evolving reusable skills and targeted practice data for recursive self-improvement in agents.

\noindent Our main contributions are as follows:
\begingroup
\setlength{\leftmargini}{1.5em}
\begin{itemize}
\setlength{\topsep}{3pt}
\setlength{\itemsep}{2pt}
\setlength{\parsep}{0pt}
\setlength{\partopsep}{0pt}
\item We identify execution trajectories as a shared foundation for both skill and data evolution, which reveal reusable procedures for skill refinement, together with core challenges and permissible variations for targeted data generation.
\item We propose V-Gym, an autonomous practice framework for agentic visual reasoning. Skill-update gains guide data selection and generation, and practice outcomes guide skill refinement.
\item Experiments show consistent gains across models and benchmarks, transfer of evolved skills across domains and models, and better skill refinement with evolved data.
\end{itemize}
\endgroup

\section{Related Work}
\label{sec:related_work}
\noindent\textbf{Distilling Skills from Experience.}
Self-evolving agents transform execution experience into reusable knowledge~\citep{gao2026surveyselfevolvingagentswhat}, including reflections, insights, workflows, and playbooks~\citep{shinn2023reflexionlanguageagentsverbal,zhao2024expelllmagentsexperiential,wang2024agentworkflowmemory,zhang2026agenticcontextengineeringevolving}. Reasoning and procedural memories further abstract past experience and support continual reuse and refinement~\citep{reasoningbank2025,memp2025,reme2025}. Experience also supports utility-aware memory retrieval and reflective prompt optimization~\citep{zhang2026memrlselfevolvingagentsruntime,agrawal2026gepareflectivepromptevolution}. Explicit skill libraries capture executable procedures~\citep{wang2023voyageropenendedembodiedagent,zheng2025skillweaverwebagentsselfimprove}, while trajectory analysis supports transferable skill extraction and revision~\citep{ni2026trace2skilldistilltrajectorylocallessons}. Candidate skill updates can be checked through student reruns or held-out validation~\citep{skillkd2026,yang2026skilloptexecutivestrategyselfevolving}. For multimodal agents, experience is organized into task-level skills, clustered knowledge, or visual procedural memories~\citep{jiang2026xskillcontinuallearningexperience,xiong2026aceskillbootstrappingmultimodalagents,skilllens2026}.

\noindent\textbf{Constructing Data from Experience.}
Agentic synthesis constructs planning and tool-use data through task generation, simulated interaction, and verification~\citep{zeng2026makesgoodagenticdata,hu2025agentgenenhancingplanningabilities,liu2025toolacewinningpointsllm,prabhakar2025apigenmtagenticpipelinemultiturn}, with multimodal extensions that diversify image-text instructions~\citep{luo2024mmevolempoweringmultimodallarge}. Beyond these synthesis pipelines, execution experience informs both task construction and training supervision: environment exploration yields executable tasks~\citep{agentevolver2025}, while multimodal tool exploration produces step-wise preference data~\citep{sport2025}. Adaptive generation uses policy rollouts, success estimates, and failure diagnoses to address evolving learning needs~\citep{huang2026onpolicydataevolutionvisualnative,wu2026envsforgefrontieroptimizedrewardgroundedenvironment,wei2026diagevodiagnosisguidedselfevolutionhierarchical}. HardGen and VISA extend this feedback to difficult tool-use samples and multimodal instruction synthesis, respectively, using failure-informed API graphs and persistent feedback from verifiers and target models~\citep{hardgen2026,visa2026}.

\noindent\textbf{Co-evolution for Agent Self-Improvement.}
Co-evolution couples updates to multiple learning components. Self-play links task proposal with solver learning~\citep{absolutezero2025,rzero2025,xia2025agent0unleashingselfevolvingagents}, including visual question generation and active image selection~\citep{visplay2025,activezero2026}. Other methods couple policies with rewards and environments~\citep{wang2026rlanythingforgeenvironmentpolicy} or tool-grounded verification~\citep{agent0vl2025}. Skills can evolve alongside policy learning~\citep{skillrl2026} or be internalized through reinforcement learning~\citep{lu2026skill0incontextagenticreinforcement}. Co-evolution also encompasses skills and tools~\citep{skillsmith2026}, skill generation and verification~\citep{zhang2026coevoskillsselfevolvingagentskills}, and solver and rubric-generator skills~\citep{decoevo2026}. Skill Self-Play and SESA connect evolving skills or procedural memory with task generation and solver training~\citep{huang2026skillselfplaypushingfrontier,fu2026selfplaymeetsskillevolution}. V-Gym studies skill--data co-evolution with a fixed multimodal solver: shared execution trajectories inform both validated skill revisions and targeted practice construction, while attributed skill-update gains guide data selection.

\begin{figure}[t]
    \centering
    \includegraphics[width=\linewidth]{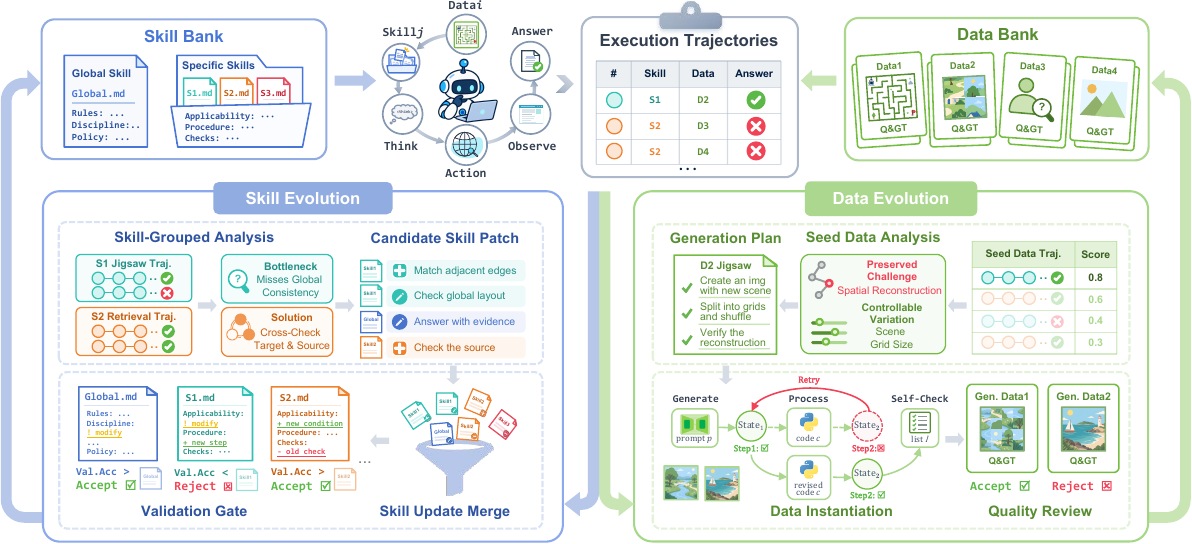}
\caption{\textbf{Overview of V-Gym.}
The agent uses the global skill and at most one routed task-specific skill in tool-assisted practice, recording trajectories and feedback.
\textit{Skill evolution} groups trajectories, consolidates patches by object, and accepts updates with positive independent validation gains.
\textit{Data evolution} ranks seeds by utility and exploration, then admits targeted generated data that pass self-checks.
The updated banks support later practice.}
    \label{fig:overview}
\end{figure}

\section{Method}
\label{sec:method}
As illustrated in Figure~\ref{fig:overview}, V-Gym improves a multimodal agent by co-evolving its skills and practice data. We first introduce the problem setting in Section~\ref{sec:problem}, then describe skill evolution and data evolution in Sections~\ref{sec:skill_evolution} and~\ref{sec:data_evolution}, respectively.

\begin{wrapfigure}{R}{0.50\textwidth}
\begin{minipage}{\linewidth}
\setlength{\intextsep}{0pt}
\begin{algorithm}[H]
\caption{V-Gym}
\label{alg:coevolution_overview}
\footnotesize
\renewcommand{\algorithmicindent}{1em}
\setboolean{ALC@noend}{true}
\begin{algorithmic}
\REQUIRE Agent $\pi_\theta$, Data Bank $D$, Validation Pool $\mathcal V$, epochs $E$, practice size $M$, rollouts $N$, budget $\rho$, exploration $\beta$, discount $\gamma$
\STATE $S\gets\varnothing$
\STATE $(A_d,n_d)\gets(0,0)$ for all $d\in D$
\FOR{$e=1,\ldots,E$}
  \STATE $\mathcal T_e\gets\textsc{Select}(D,M;\sigma_d)$
  \STATE $\mathcal U_e^+\gets\varnothing$
  \FOR{each batch $B_{e,b}$ of $\mathcal T_e$}
    \STATE $S_{e,b}\gets S$
    \STATE $\textsc{Practice}(\pi_\theta,S_{e,b},B_{e,b},N)$
    \STATE $\{(\delta_o,\mathcal I_o)\}\gets\textsc{Refine}(S_{e,b},B_{e,b})$
    \STATE $\{R_o\}\gets\textsc{Validate}(S_{e,b},\{(\delta_o,\mathcal I_o)\},\mathcal V)$
    \STATE $S\gets\textsc{Merge}(S_{e,b},\{\delta_o:R_o>0\})$
    \STATE $\mathcal U_e^+\gets\mathcal U_e^+\cup\{(e,b,o):R_o>0\}$
  \ENDFOR
  \FOR{each $d\in\mathcal T_e$ with complete feedback}
    \STATE Compute $U_d$ by Eq.~\eqref{eq:practice_utility}
    \STATE $(A_d,n_d)\gets(A_d+U_d,n_d+1)$
  \ENDFOR
  \IF{$e<E$}
    \STATE Rank eligible $d\in\mathcal T_e$ by $\sigma_d$ \eqref{eq:selection_score}
    \STATE $D\gets\textsc{GenerateChecked}(D,\mathcal T_e;\lfloor\rho M\rfloor)$
    \STATE $(A_d,n_d)\gets(0,0)$ for newly admitted $d$
    \STATE $(A_d,n_d)\gets(\gamma A_d,\gamma n_d)$ for all $d\in D$
  \ENDIF
\ENDFOR
\RETURN $S$, frozen for inference
\end{algorithmic}
\end{algorithm}
\end{minipage}
\end{wrapfigure}

\subsection{Problem Setting}
\label{sec:problem}

We use a fixed-parameter multimodal tool-using agent $\pi_\theta$ with skill bank $S$ and data bank $D$. Skill evolution maximizes the task-solving improvement brought by skill updates, while data evolution maximizes the value of practice data in supporting further skill improvement. Let $R(S';S)$ denote the skill-update reward and $U(D';S)$ denote the expected usefulness of a data bank $D$ with skill bank $S$, informed by the same evidence. Given the current skill bank $S_k$ and data bank $D$, we consider candidate expansions $D'\supseteq D$ within a fixed generation budget and express these coupled goals as:

\begin{equation}
\label{eq:coevolution_objective}
S^+=\operatorname*{arg\,max}_{S'}\;
\mathbb E_{\substack{D^+}}
\bigl[R(S';S)\bigr]
, \quad
\text{where}\
D^+=\operatorname*{arg\,max}_{D'\supseteq D}\quad
\mathbb E_{\substack{D}}
\bigl[U(D';S)\bigr],
\end{equation}
where $S^+$ and $D^+$ are evolved skill and data banks. Both evolution processes use practice trajectories collected under the current $S$ and $D$. Specifically, each instance $d=(I,q,y)$ contains images $I$, a question $q$, and reference information $y$. At each iteration, we select $d\in D$ and task-dependent skill $s\in S$. The agent produces a tool-assisted reasoning trajectory $t\sim\pi_\theta(\cdot\mid I,q,s)$ and receives answer-evaluation feedback. We estimate skill-update rewards from trajectories through comparative validation in Section~\ref{sec:skill_evolution} and attribute these rewards to practice instances to guide data evolution in Section~\ref{sec:data_evolution}. Algorithm~\ref{alg:coevolution_overview} summarizes the full co-evolution process.

\subsection{Skill evolution}
\label{sec:skill_evolution}

Skill evolution grounds revisions in execution evidence and retains them through comparative validation. The skill bank combines a global skill $S_g$ with task-specific skills $\{S^{(k)}\}_{k=1}^{K}$. The applicability metadata of task-specific skills automatically form a router document. When solving a task, the agent consults this document and selects at most one applicable specific skill. Writing its choice as $k_d\in\{0,\ldots,K\}$, the selected context is $s=S_g\oplus S^{(k_d)}$, where $\oplus$ concatenates documents and $S^{(0)}=\varnothing$ denotes no match.

\paragraph{Trajectory-Grounded Skill Refinement.}
Starting from an empty skill bank and initial data bank $D$, we run $E$ epochs and update $D$ in place. At epoch $e$, a fixed practice set $\mathcal T_e\subseteq D$ of $M$ instances is partitioned into batches. For batch $B_{e,b}$, \textsc{Practice} runs $N$ rollouts per instance under the batch-start bank $S_{e,b}$, retaining the resulting $N|B_{e,b}|$ trajectories and evaluation feedback for skill refinement and data generation. \textsc{Refine} uses these recorded trajectories, grouping them by routed specific skill or, when unmatched, the semantic similarity of their source instances. Within each group, reflection compares successful and failed trajectories to identify reusable workflows, missing visual evidence, reasoning errors, and inappropriate applicability conditions. It proposes skill patches that either edit existing global or specific skills or create new skills with complete procedures and metadata.

Skill patches from all groups targeting the same object $o$ are deduplicated and merged into one candidate update $\delta_o$. For example, patches to a specific skill's procedure and applicability conditions are merged into a single update to that skill. Each update retains the source instance set $\mathcal I_o$ of distinct contributors to these groups for utility attribution.

\paragraph{Comparative Validation.}
Let $r(d;S)$ denote the mean score on instance $d$ across rollouts under bank $S$. \textsc{Validate} instantiates the skill reward $R$ as the validation improvement of a candidate skill update. Its contributing instances are used to retrieve a nonempty, deduplicated panel $\mathcal V_o$ from a separate validation data pool $\mathcal V$. Each candidate skill update is applied independently to the same batch-start bank $S_{e,b}$ to obtain $\widetilde{S}_{e,b}^{\,o}$. We compare the agent equipped with $\widetilde{S}_{e,b}^{\,o}$ and the agent equipped with $S_{e,b}$ on the candidate's validation panel using fresh rollouts under matched settings, obtaining the skill-update reward as follows:

\begin{equation}
\label{eq:skill_reward}
R_o=\frac{1}{|\mathcal V_o|}\sum_{d\in\mathcal V_o}
\bigl[r(d;\widetilde{S}_{e,b}^{\,o})-r(d;S_{e,b})\bigr].
\end{equation}

\textsc{Merge} accepts all candidate object-level skill updates with $R_o>0$ and merges them into $S_{e,b}$ to form $S_{e,b+1}$, with the router refreshed accordingly. Each accepted update retains its reward and contributing instances for data-utility attribution.

\subsection{Data evolution}
\label{sec:data_evolution}

After all batches of an epoch, data evolution attributes skill-update rewards to the practiced instances, updating utility statistics for the current data bank $D$. We balance utility and exploration through UCB-style seed selection, then use the selected seeds' execution trajectories to specify new practice data that expand and evolve the data bank.

\paragraph{Utility--Exploration Selection.}
Each accepted update event $u=(e,b,o)\in\mathcal U_e^{+}$ retains the reward $R_u$ from Eq.~\eqref{eq:skill_reward} and its contributing instance set $\mathcal I_u$. We instantiate an instance's utility by attributing an equal share of each accepted reward to its contributors:

\begin{equation}
\label{eq:practice_utility}
U_d=\sum_{u\in\mathcal U_e^{+}:\,d\in\mathcal I_u}\frac{R_u}{|\mathcal I_u|},
\end{equation}

where $\mathcal U_e^{+}$ contains the accepted update events in epoch $e$. Attribution aggregates contributions across each instance's rollouts and skill updates. Only instances with complete trajectories, reflection, and all relevant validation outcomes receive one utility observation and qualify as seeds. A complete instance supporting no accepted update has $U_d=0$, while incomplete feedback yields no observation.

Inspired by upper confidence bound (UCB) methods, we define a selection score $\sigma_d$ from accumulated utility $A_d$ and effective practice count $n_d$:

\begin{equation}
\label{eq:selection_score}
\sigma_d=
\begin{cases}
+\infty, & n_d=0,\\[2pt]
\displaystyle\frac{A_d}{n_d}
+\beta\sqrt{\frac{2\log\bigl(1+\sum_{\bar d\in D} n_{\bar d}\bigr)}{n_d}}, & \text{otherwise},
\end{cases}
\end{equation}

for $d\in D$, where $\beta\geq0$ controls exploration. We initialize $A_d=n_d=0$ on admission and, once after complete practice in an epoch, add $U_d$ to $A_d$ and one to $n_d$. Mean utility favors instances supporting useful skill updates, while exploration revisits less-practiced instances.

For $e<E$, the generation budget $\rho\in[0,1]$ targets at most $\lfloor\rho M\rfloor$ admissions from this epoch's eligible practice seeds in descending $\sigma_d$ order.

\paragraph{Targeted Data Generation.}
The selection score $\sigma_d$ determines which practice instances are used as seeds to generate new instances, while their trajectories specify what these new instances should preserve and vary. For each seed $d$, trajectory analysis identifies the task's key challenges and data-format requirements, together with aspects that can be varied. The generation plan preserves these requirements and comparable difficulty while introducing diversity. For example, Figure~\ref{fig:trajectory} varies base tasks and scenes while retaining the low-visibility challenge.

The generator executes this plan with the enabled construction, search, and image-processing tools, derives the answer from the resulting artifact, and records supporting evidence. Model self-checks assess four requirements: reliable answers and evidence, image-question consistency, preservation of core task challenges, and absence of answer leakage or shortcuts. Candidates that still fail these checks after bounded repair are rejected. In Algorithm~\ref{alg:coevolution_overview}, \textsc{GenerateChecked} traverses the ranked eligible seeds using their recorded trajectories, stopping at the admission budget or when seeds are exhausted. Each successful seed contributes one accepted instance directly to the shared data bank $D$. Per-seed attempt limits and admission checks are detailed in Appendix~\ref{app:data_evolution_details}.

New instances start with $A_d=n_d=0$. We discount all statistics by $(A_d,n_d)\leftarrow(\gamma A_d,\gamma n_d)$, where $\gamma\in(0,1]$. For the next epoch, \textsc{Select} takes the top $M$ real and generated instances from the shared data bank by the same selection score $\sigma_d$.

\section{Experiments}
\label{sec:experiments}
This section examines overall task-solving performance, transfer across domains and backbone models, the contribution of each component, and the quality and usefulness of evolved practice data. We first describe the experimental setup, then report main results, transfer results, ablations, and further analyses.

\begin{table}[t]
\centering
\caption{Task-solving performance across three backbones and three benchmarks (\%).}
\label{tab:main_benchmark_results}
\setlength{\tabcolsep}{3pt}
\resizebox{\textwidth}{!}{%
\begin{tabular}{lcccccccc}
\toprule
\multirow{2}{*}{\textbf{Method}}
& \multicolumn{2}{c}{\textbf{TIRBench}}
& \multicolumn{2}{c}{\textbf{MMSearch-Plus}}
& \multicolumn{2}{c}{\textbf{MMBrowseComp}}
& \multicolumn{2}{c}{\textbf{Average}} \\
\cmidrule(lr){2-3}
\cmidrule(lr){4-5}
\cmidrule(lr){6-7}
\cmidrule(lr){8-9}
& Avg.@3 & Pass@3 & Avg.@3 & Pass@3
& Avg.@3 & Pass@3 & Avg.@3 & Pass@3 \\
\midrule
\multicolumn{9}{c}{\textbf{GPT-5.5}} \\
\midrule
Baseline & 40.9 & 57.7 & 20.3 & 36.0 & 15.6 & 22.7 & 25.6 & 38.8 \\
Vanilla Tools & 63.1 & 73.8 & 34.0 & 49.0 & 33.3 & 46.0 & 43.5 & 56.3 \\
XSkill~\citep{jiang2026xskillcontinuallearningexperience} & 64.7 & 77.4 & 34.3 & 54.0 & 36.2 & 54.7 & 45.1 & 62.0 \\
Ace-Skill~\citep{xiong2026aceskillbootstrappingmultimodalagents} & 64.7 & 78.7 & 36.0 & 50.0 & 36.7 & 56.0 & 45.8 & 61.6 \\
SkillOPT~\citep{yang2026skilloptexecutivestrategyselfevolving} & 66.2 & 78.2 & 36.3 & 51.0 & 35.3 & 55.3 & 45.9 & 61.5 \\
\rowcolor[HTML]{E7ECF8}
V-Gym (Ours) & \textbf{71.1} & \textbf{81.5} & \textbf{40.0} & \textbf{57.0} & \textbf{38.7} & \textbf{57.3} & \textbf{49.9} & \textbf{65.3} \\
\midrule
\multicolumn{9}{c}{\textbf{Gemini-3.5-Flash}} \\
\midrule
Baseline & 43.6 & 60.3 & 30.3 & 39.0 & 12.9 & 22.0 & 28.9 & 40.4 \\
Vanilla Tools & 47.7 & 68.2 & 46.3 & 61.0 & 18.9 & 30.7 & 37.6 & 53.3 \\
\rowcolor[HTML]{E7ECF8}
V-Gym (Ours) & \textbf{55.7} & \textbf{72.1} & \textbf{56.0} & \textbf{70.0} & \textbf{26.7} & \textbf{37.3} & \textbf{46.1} & \textbf{59.8} \\
\midrule
\multicolumn{9}{c}{\textbf{Qwen-3.7-Flash}} \\
\midrule
Baseline & 27.9 & 37.4 & 14.3 & 20.0 & 7.8 & 16.7 & 16.7 & 24.7 \\
Vanilla Tools & 58.6 & 73.1 & 32.7 & 49.0 & 13.6 & 26.7 & 35.0 & 49.6 \\
\rowcolor[HTML]{E7ECF8}
V-Gym (Ours) & \textbf{69.9} & \textbf{78.5} & \textbf{39.0} & \textbf{57.0} & \textbf{17.6} & \textbf{34.0} & \textbf{42.2} & \textbf{56.5} \\
\bottomrule
\end{tabular}%
}
\end{table}

\subsection{Experimental Setup}

Our main evaluation uses TIRBench~\citep{li2025tirbenchcomprehensivebenchmarkagentic},
MMSearch-Plus~\citep{tao2026mmsearchplusbenchmarkingprovenanceawaresearch}, and
MMBrowseComp~\citep{li2026mmbrowsecompcomprehensivebenchmarkmultimodal}, with
GPT-5.5~\citep{openai2026gpt55}, Gemini-3.5-Flash~\citep{kavukcuoglu2026gemini35}, and
Qwen-3.7-Flash~\citep{alibaba2026qwen37flash} as backbones.
We compare V-Gym with direct chain-of-thought (CoT) reasoning (Baseline), tool-assisted solving (Vanilla Tools),
and existing multimodal skill-evolution methods, including XSkill~\citep{jiang2026xskillcontinuallearningexperience},
Ace-Skill~\citep{xiong2026aceskillbootstrappingmultimodalagents}, and
SkillOPT~\citep{yang2026skilloptexecutivestrategyselfevolving} using GPT-5.5. All skill-evolution methods share the same training rollout budget. We follow each method's original settings except for the training and validation controls specified in Appendix~\ref{app:comparison_protocol}. V-Gym uses 2 practice epochs by default, a batch size of 5, and a generation budget for data evolution of $\rho=0.6$. At test time, we run three rollouts per data instance and report Avg.@3 and Pass@3 accuracy. Further experimental details are in Appendix~\ref{app:experimental_settings}. To characterize test-time variability, we also report the mean per-benchmark standard deviation of single-run accuracy across three test runs in Appendix~\ref{app:test_time_variability}.

\subsection{Main Results}

We evaluate whether V-Gym's autonomous practice produces skills that improve agentic visual reasoning beyond tool access and existing skill-evolution methods.
As shown in Table~\ref{tab:main_benchmark_results}, V-Gym achieves the highest Avg.@3 and Pass@3 among the compared methods with GPT-5.5 on all three benchmarks.
Relative to the strongest competing result for each metric, Avg.@3 improves by 4.9, 3.7, and 2.0 points on TIRBench, MMSearch-Plus, and MMBrowseComp, respectively, with corresponding Pass@3 gains of 2.8, 3.0, and 1.3 points.
These consistent gains show that the evolved skill bank improves performance across varied visual reasoning tasks, including multimodal tool use, search, and browsing.

The same consistent improvements over Vanilla Tools with Gemini-3.5-Flash and Qwen-3.7-Flash, across all three benchmarks and both metrics, further demonstrate V-Gym's ability to enhance agentic visual reasoning across different models and datasets. 

\begin{table}[t]
\centering
\caption{Out-of-distribution transfer evaluation results (\%). Skills acquired from the source domain or model are directly applied to the target domain or model for evaluation.}
\label{tab:transfer}
\label{tab:benchmark_transfer}
\label{tab:model_transfer}
\begin{minipage}[t]{0.38\textwidth}
\vspace{0pt}
\centering
\fontsize{8}{9.5}\selectfont
\setlength{\tabcolsep}{1.5pt}
\renewcommand{\arraystretch}{1.15}
(a) Cross-Domain Transfer\par\vspace{3pt}
\begin{tabular}{p{\dimexpr0.38\linewidth-2\tabcolsep\relax}*{2}{>{\centering\arraybackslash}p{\dimexpr0.31\linewidth-2\tabcolsep\relax}}}
\toprule
\multirow{2}{*}{\textbf{Method}} & \multicolumn{2}{c}{\textbf{Target Benchmark}}\\
\cmidrule(lr){2-3}
 & Avg.@3 & Pass@3 \\
\midrule
\multicolumn{3}{c}{\textbf{TIRBench $\rightarrow$ VisualToolBench}} \\
\midrule
Vanilla Tools & 39.6 & 58.7 \\
\rowcolor[HTML]{E7ECF8}
V-Gym (Ours) & 46.2~\textcolor[HTML]{228B22}{{($\uparrow$6.6)}} & 66.7~\textcolor[HTML]{228B22}{{($\uparrow$8.0)}} \\
\midrule
\multicolumn{3}{c}{\textbf{MMSearch-Plus $\rightarrow$ AgentVista}} \\
\midrule
Vanilla Tools & 31.3 & 41.6 \\
\rowcolor[HTML]{E7ECF8}
V-Gym (Ours) & 35.4~\textcolor[HTML]{228B22}{{($\uparrow$4.1)}} & 46.9~\textcolor[HTML]{228B22}{{($\uparrow$5.3)}} \\
\bottomrule
\end{tabular}
\end{minipage}\hfill
\begin{minipage}[t]{0.60\textwidth}
\vspace{0pt}
\centering
\fontsize{8}{9.5}\selectfont
\setlength{\tabcolsep}{1.5pt}
\renewcommand{\arraystretch}{1.15}
(b) Cross-Model Transfer\par\vspace{3pt}
\begin{tabular}{p{\dimexpr0.22\linewidth-2\tabcolsep\relax}*{4}{>{\centering\arraybackslash}p{\dimexpr0.195\linewidth-2\tabcolsep\relax}}}
\toprule
\multirow{2}{*}{\textbf{Method}} & \multicolumn{2}{c}{\textbf{TIRBench}} & \multicolumn{2}{c}{\textbf{MMSearch-Plus}} \\
\cmidrule(lr){2-3}\cmidrule(lr){4-5}
& Avg.@3 & Pass@3 & Avg.@3 & Pass@3 \\
\midrule
\multicolumn{5}{c}{\textbf{GPT-5.5 $\rightarrow$ GPT-5.4-mini}} \\
\midrule
Vanilla Tools & 36.7 & 58.5 & 15.7 & 29.0 \\
\rowcolor[HTML]{E7ECF8}
V-Gym (Ours) & 43.4~\textcolor[HTML]{228B22}{{($\uparrow$6.7)}} & 70.0~\textcolor[HTML]{228B22}{{($\uparrow$11.5)}} & 21.3~\textcolor[HTML]{228B22}{{($\uparrow$5.6)}} & 38.0~\textcolor[HTML]{228B22}{{($\uparrow$9.0)}} \\
\midrule
\multicolumn{5}{c}{\textbf{GPT-5.5 $\rightarrow$ Qwen-3.5-Flash}} \\
\midrule
Vanilla Tools & 42.1 & 62.1 & 23.7 & 38.0 \\
\rowcolor[HTML]{E7ECF8}
V-Gym (Ours) & 46.2~\textcolor[HTML]{228B22}{{($\uparrow$4.1)}} & 64.4~\textcolor[HTML]{228B22}{{($\uparrow$2.3)}} & 25.3~\textcolor[HTML]{228B22}{{($\uparrow$1.6)}} & 40.0~\textcolor[HTML]{228B22}{{($\uparrow$2.0)}} \\
\bottomrule
\end{tabular}
\end{minipage}
\end{table}

\subsection{Out-of-Distribution Transfer}
\label{sec:ood_transfer}

\paragraph{Domain transfer.}
We test whether V-Gym's evolved skill bank generalizes to unseen domains.
In Table~\ref{tab:benchmark_transfer}(a), skill banks evolved with GPT-5.5 on TIRBench and MMSearch-Plus are directly transferred to VisualToolBench~\citep{guo2025seeingevaluatingmultimodalllms} and AgentVista~\citep{su2026agentvistaevaluatingmultimodalagents}, respectively, without any further skill updates.
Compared with Vanilla Tools, transfer raises Avg.@3 from 39.6 to 46.2 on VisualToolBench and from 31.3 to 35.4 on AgentVista, with gains in Pass@3 on both targets as well.
These improvements show that the skills acquired through V-Gym's co-evolved practice remain effective beyond their source domains, supporting their generality across domains.

\paragraph{Model transfer.}
We further evaluate model transfer to test whether the evolved skill bank can improve less capable agents.
Table~\ref{tab:model_transfer}(b) transfers skills acquired with GPT-5.5 without further updates to the same-family
GPT-5.4-mini~\citep{openai2026gpt54mini} or the cross-family
Qwen-3.5-Flash~\citep{alibaba2026qwen35flash}.
Both target models improve over Vanilla Tools on both benchmarks and both metrics, for example, GPT-5.4-mini gains 6.7 and 5.6 Avg.@3 points on TIRBench and MMSearch-Plus, respectively.
These gains demonstrate that V-Gym acquires effective skills that transfer to less capable models within and across model families.

\begin{figure}[t]
\centering
\begin{minipage}[t]{0.50\linewidth}
\vspace{0pt}
\centering
\makeatletter\def\@captype{table}\makeatother
\caption{Ablation on core component (\%).}
\label{tab:component_ablation}
\fontsize{7.5}{9}\selectfont
\setlength{\tabcolsep}{2pt}
\renewcommand{\arraystretch}{1.12}
\begin{tabular}{p{\dimexpr0.45\linewidth-2\tabcolsep\relax}*{2}{>{\centering\arraybackslash}p{\dimexpr0.265\linewidth-2\tabcolsep\relax}}}
\toprule
\multirow{2}{*}{\textbf{Configuration}} & \multicolumn{2}{c}{\textbf{TIRBench}} \\
\cmidrule(lr){2-3}
 & \textbf{Avg.@3} & \textbf{Pass@3} \\
\midrule
 w/ Tools & 63.1 & 73.8 \\
 w/ Skill Evo. & 68.5 & 79.5 \\
\rowcolor[HTML]{E7ECF8}
 w/ Skill-Data Co-Evo. & \textbf{71.1} & \textbf{81.5} \\
\midrule
\multicolumn{3}{c}{\textit{Skill Evolution Ablation}} \\
\midrule
w/o Hierarchical Bank & 69.3~\textcolor[HTML]{CC0000}{{($\downarrow$1.8)}} & 79.2~\textcolor[HTML]{CC0000}{{($\downarrow$2.3)}} \\
w/o Validation Gate & 65.8~\textcolor[HTML]{CC0000}{{($\downarrow$5.3)}} & 78.7~\textcolor[HTML]{CC0000}{{($\downarrow$2.8)}} \\
\midrule
\multicolumn{3}{c}{\textit{Data Evolution Ablation}} \\
\midrule
w/o UCB Sampling & 69.1~\textcolor[HTML]{CC0000}{{($\downarrow$2.0)}} & 79.7~\textcolor[HTML]{CC0000}{{($\downarrow$1.8)}} \\
w/o Targeted Spec. & 68.9~\textcolor[HTML]{CC0000}{{($\downarrow$2.2)}} & 80.0~\textcolor[HTML]{CC0000}{{($\downarrow$1.5)}} \\
\bottomrule
\end{tabular}
\end{minipage}\hfill
\begin{minipage}[t]{0.48\linewidth}
\vspace{6.6pt}
\centering
\includegraphics[width=0.98\linewidth]{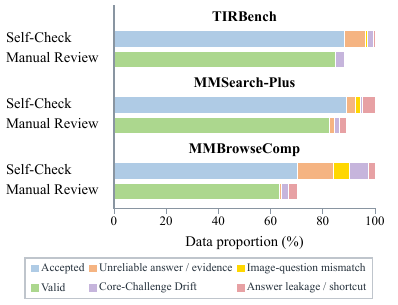}\par
\nointerlineskip
\setlength{\abovecaptionskip}{0pt}
\caption{Quality distribution of generated data based on self-checks and manual review.}
\label{fig:data_distribution}
\end{minipage}
\end{figure}

\subsection{Ablation Study}

\paragraph{Contribution of each component.}
To assess the contribution of each component, we conduct ablations at epoch 2 on TIRBench.
As shown in Table~\ref{tab:component_ablation}, adding skill evolution to tool use improves Avg.@3/Pass@3 from 63.1/73.8 to 68.5/79.5, and co-evolving practice data further raises performance to 71.1/81.5.
Without the hierarchical bank, all guidance is maintained in a single skill document without separate global and task-specific layers or a router document.
Removing the validation gate retains validation scores for utility estimation but accepts updates without filtering by gain.
Both changes reduce performance, with removal of the validation gate causing the largest Avg.@3 drop, highlighting the value of filtering skill updates by measured improvement.

Within data evolution, replacing UCB-style sampling with random selection for both generation seeds and the next epoch's practice data also lowers performance.
Without targeted specification, the generator receives only the seed image-text pair and generates directly, bypassing trajectory analysis while retaining UCB-style sampling and the remaining settings.
This variant recovers little of the full method's gain over skill evolution alone, supporting the value of trajectory-derived generation guidance.
Removing any component lowers both metrics, and the full method performs best, supporting the contributions of the individual components and their combined design.

\subsection{Further Analysis}
\label{sec:data_evolution_analysis}

\paragraph{Evolved-data quality.}
\label{sec:data_quality}
To assess whether evolved data provide valid practice inputs while preserving the core task challenges, we examine generation self-checks and manually review accepted samples.
Figure~\ref{fig:data_distribution} shows model acceptance rates of 70.4-88.8\% across the three benchmarks, with unreliable answers or evidence and answer leakage or shortcuts as the most common rejection reasons.
This distribution shows how self-checks screen candidate defects before admission.
For an independent assessment, we randomly sample 100 model-accepted instances per benchmark and conduct three independent manual reviews of the same samples for validity, preservation of core task challenges, and comparable difficulty.
The proportions satisfying these requirements are 96\%, 93\%, and 90\% on TIRBench, MMSearch-Plus, and MMBrowseComp, respectively, with a mean pairwise agreement of 93.56\% across reviews.
These results support the reliability of V-Gym's admitted practice data and their preservation of the core task challenges.
Appendix~\ref{app:human_review} details the review protocol and~disagreement~resolution.

\begin{figure}[!t]
\centering
\edef\vgympanelheight{\the\dimexpr0.37\linewidth\relax}
\begin{minipage}[t]{0.344\linewidth}
\centering\vspace{0pt}
\includegraphics[height=\vgympanelheight,trim=9bp 5bp 7bp 10bp,clip]{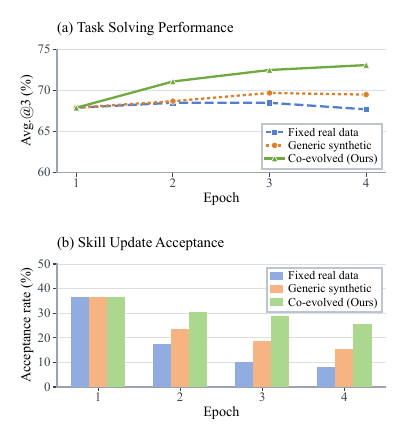}
\end{minipage}\hfill
\begin{minipage}[t][\vgympanelheight][s]{0.636\linewidth}
\centering\vspace{0pt}
{\fontsize{5.4}{6.5}\selectfont(c) Examples of Evolved Skill\par}
\vfill
\includegraphics[height=\dimexpr\vgympanelheight-10pt\relax,pagebox=cropbox]{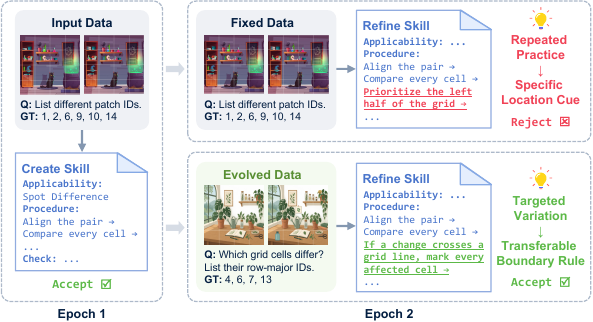}
\end{minipage}
\caption{Skill evolution with different practice data on TIRBench with GPT-5.5. We compare (a) task-solving performance, (b) skill-update acceptance across epochs, and (c) example skills evolved with fixed data and our co-evolved data. Epoch 1 is shared warm-up.}
\label{fig:data_evolution}
\end{figure}

\paragraph{Effectiveness of continued co-evolution.}
To test whether adapting practice data sustains skill evolution over time, we compare fixed-real data, generic synthetic data, and online co-evolution over four epochs with shared initial warm-up.
The two generation strategies use the same budget. Generic synthesis uses randomly selected seeds corresponding to 60\% of the practice budget, then mixes 60\% synthetic and 40\% real data for subsequent practice.
Figure~\ref{fig:data_evolution}(a) shows that fixed-real practice plateaus and then declines, while generic synthesis improves modestly before leveling off.
Co-evolution continues to improve throughout the observed epochs, finishing 3.6 and 5.4 Avg.@3 points above generic synthetic and fixed-real data, respectively.
This trend demonstrates the sustained benefit of adapting practice data as the skill bank evolves.

\paragraph{Skill-update acceptance.}
To examine whether this performance trend is accompanied by continued useful skill revisions, Figure~\ref{fig:data_evolution}(b) tracks the fraction of consolidated candidate updates accepted per epoch.
An update is counted as accepted only when it yields a positive validation gain over the corresponding batch-start skill bank.
Acceptance declines for all strategies, but co-evolution maintains the highest rate after the shared warm-up, ending at 25.6\% compared with 15.4\% for generic synthesis and 8.2\% for fixed-real practice.
This late-stage gap indicates that co-evolution continues to produce a larger proportion of beneficial updates even as successful revisions become less frequent overall.
Together, the two trends show that V-Gym sustains a larger share of successful skill revisions alongside continued task-solving gains.

\paragraph{Case studies.}
\label{sec:case_studies}
To illustrate how practice data shape skill refinement, Figure~\ref{fig:data_evolution}(c) compares skills evolved with fixed and co-evolved data. Co-evolved practice yields a more general refinement that passes validation, while repeated fixed data produce an instance-specific revision that is rejected. Specifically, the fixed-data revision prioritizes the left half of the grid, whereas the co-evolved revision introduces a rule to mark every affected cell when a change crosses a grid boundary. This example illustrates how V-Gym's evolving practice data can support reusable skill improvements during continued practice.

\section{Limitations}
\label{sec:limitations}
V-Gym is an initial exploration of skill-data co-evolution for agentic visual reasoning. Its current data evolution process remains limited in its ability to construct reliable practice for highly complex real-world multimodal tasks and long videos. Nevertheless, transfer to AgentVista shows that skills acquired within the current framework can already support reasoning in a challenging target domain, while advances in generation tools may enable broader coverage of task types and longer temporal horizons. Further scaling is also constrained by the substantial computational costs of repeated agent rollouts, skill-update validation, and data generation, which limit the size of our training and validation sets. Larger and more diverse high-quality practice datasets, together with broader validation coverage, may further improve skill acquisition and update selection. Exploring these potential benefits at scale remains an important direction for future work.

\section{Conclusion}
\label{sec:conclusion}
We propose V-Gym, an autonomous practice framework for improving agentic visual reasoning through skill-data co-evolution with fixed model parameters. Execution trajectories provide shared evidence for refining reusable skills and specifying targeted, diverse practice, while validation feedback guides subsequent updates and data selection. Extensive evaluation results show improvements over the compared baselines across diverse visual reasoning benchmarks, as well as generalization to unseen domains and transfer to weaker models. These findings support the potential of coordinating what an agent learns from experience with what it practices next, making skill refinement and adaptive practice complementary components of agent self-improvement.

\FloatBarrier
\subsection*{AI use statement}

For manuscript preparation, we used generative AI tools only for language polishing and the retrieval and discovery of relevant literature. As part of V-Gym, we also used generative AI tools to construct synthetic multimodal practice data. Generated data undergo model self-checks for answer reliability, image-question consistency, preservation of core task challenges, and absence of answer leakage. We further assessed generated-data quality through independent human review of sampled instances, as described in Appendix~\ref{app:human_review}. We have reviewed and revised the AI-assisted text and checked the retrieved references for accuracy and relevance. We take responsibility for the final content of this work, including its text, citations, claims, and artifacts produced with the aid of generative AI.

\subsection*{Reproducibility statement}

To support reproducibility, we describe V-Gym in Section~\ref{sec:method} and provide its detailed algorithm, skill-update validation procedure, and data-evolution rules in Appendix~\ref{app:algorithm}. Appendix~\ref{app:experimental_settings} documents dataset partitions, sampling procedures, tool availability, model configurations, and hyperparameters. Appendix~\ref{app:prompts} provides prompt templates for the execution and evolution stages, while Appendix~\ref{app:skill_examples} presents examples of the learned skill bank. The evaluation protocol and test-time variability are described in Section~\ref{sec:experiments} and Appendix~\ref{app:test_time_variability}, respectively. Appendix~\ref{app:human_review} details the sampling, assessment criteria, and disagreement-resolution procedure used for human review of generated data.

\bibliography{references}
\bibliographystyle{vgym}
\clearpage
\appendix
\section{Method Details}
\label{app:algorithm}

\subsection{Detailed Co-Evolution Algorithm}
\label{app:full_algorithm}

Algorithm~\ref{alg:coevolution} expands the co-evolution process in Algorithm~\ref{alg:coevolution_overview}. Within each epoch, practice and skill refinement proceed batch by batch, while utility accounting follows after all batches, and data generation runs only when $e<E$. The resulting data bank supports selection for the next epoch.

\begin{algorithm}[H]
\caption{Detailed V-Gym co-evolution}
\label{alg:coevolution}
\small
\begin{algorithmic}[1]
\REQUIRE Agent $\pi_\theta$, Data Bank $D$, Validation Pool $\mathcal V$, epochs $E$, practice size $M$, rollouts $N$, generation budget $\rho$, exploration coefficient $\beta$, discount $\gamma$
\STATE Initialize $S\gets\varnothing$ and $A_d=n_d=0$ for $d\in D$
\FOR{$e=1,\ldots,E$}
  \STATE Select the $M$ instances with the highest selection scores $\sigma_d$ from $D$ as $\mathcal T_e$, breaking ties randomly and shuffling the execution order
  \STATE Partition $\mathcal T_e$ into batches $B_{e,b}$ and initialize accepted update events $\mathcal U_e^+\gets\varnothing$
  \STATE \textbf{Skill evolution}
  \FOR{each batch $B_{e,b}$}
    \STATE Fix the batch-start bank $S_{e,b}\gets S$ for practice and candidate validation
    \STATE Collect $N|B_{e,b}|$ trajectories with evaluation feedback using $\pi_\theta$ and $S_{e,b}$
    \STATE Group trajectories by routed skill or, if unmatched, source-instance similarity, then reflect and propose skill patches
    \STATE Deduplicate and merge patches by object $o$ into $\delta_o$ and retain distinct contributors $\mathcal I_o$
    \FOR{each candidate skill update $\delta_o$}
      \STATE Retrieve a nonempty, deduplicated panel $\mathcal V_o\subseteq\mathcal V$ using $\mathcal I_o$
      \STATE Apply $\delta_o$ independently to $S_{e,b}$ to obtain $\widetilde S_{e,b}^{\,o}$
      \STATE Compare $\widetilde S_{e,b}^{\,o}$ and $S_{e,b}$ on $\mathcal V_o$ using fresh rollouts under matched settings and compute $R_o$ by Eq.~\eqref{eq:skill_reward}
    \ENDFOR
    \STATE Merge updates with $R_o>0$ into $S_{e,b}$ to form $S_{e,b+1}$, then set $S\gets S_{e,b+1}$ and refresh the router
    \STATE Record accepted events $u=(e,b,o)$ in $\mathcal U_e^+$ with rewards $R_u$ and contributors $\mathcal I_u$
  \ENDFOR
  \STATE \textbf{Data evolution}
  \FOR{each $d\in\mathcal T_e$ with complete practice feedback}
    \STATE Compute $U_d$ by Eq.~\eqref{eq:practice_utility} and update $A_d\gets A_d+U_d$ and $n_d\gets n_d+1$ once
  \ENDFOR
  \IF{$e<E$}
    \STATE Rank complete-feedback seeds $d\in\mathcal T_e$ by $\sigma_d$ in Eq.~\eqref{eq:selection_score}
    \FOR{each ranked seed until $\lfloor\rho M\rfloor$ new instances are admitted or seeds are exhausted}
      \STATE Derive a generation plan from the seed's execution trajectories
      \STATE Generate and self-check candidates, stopping at the first accepted instance or five failed attempts
      \STATE Add any accepted instance to $D$ and initialize its $A_d=n_d=0$
    \ENDFOR
    \STATE Discount $(A_d,n_d)\gets(\gamma A_d,\gamma n_d)$ for every $d\in D$
  \ENDIF
\ENDFOR
\RETURN Final skill bank $S$, frozen for inference
\end{algorithmic}
\end{algorithm}

\subsection{Skill Evolution Details}
\label{app:skill_evolution_details}

\paragraph{Skill patches and update objects.}
Reflection over each trajectory group proposes patches that edit existing global or task-specific skills or create new skills with complete procedures and applicability metadata. A patch specifies its target skill, operation, text location, and revised content. Patches from all groups that target the same object $o$ are deduplicated and merged into one candidate update $\delta_o$ before validation. For example, edits to a specific skill's procedure and applicability conditions form one update. Its contributor set $\mathcal I_o$ contains the distinct source instances of the contributing groups, so multiple trajectories from an instance do not multiply its share of that update's reward.

\paragraph{Comparative validation.}
The contributing instances query an embedding index over the separate validation data pool $\mathcal V$. Cosine similarity identifies relevant instances, producing a nonempty, deduplicated panel $\mathcal V_o$ subject to the retrieval and size limits in Table~\ref{tab:appendix_hyperparameters}. Each candidate skill update is applied independently to the same batch-start bank $S_{e,b}$, which remains fixed throughout these comparisons. Agents equipped with the candidate bank $\widetilde S_{e,b}^{\,o}$ and the original bank solve the same panel using fresh rollouts under matched settings. The mean score difference in Eq.~\eqref{eq:skill_reward} gives $R_o$. After these comparisons, all updates with $R_o>0$ are merged into $S_{e,b}$ to form $S_{e,b+1}$, and the router is refreshed from the task-specific skills' applicability metadata. The validation data pool excludes practice and held-out test instances.

\subsection{Data Evolution Details}
\label{app:data_evolution_details}

\paragraph{Utility--exploration selection.}
Only instances in the current practice set $\mathcal T_e$ with complete trajectories, reflection, and all relevant validation outcomes receive a utility observation and qualify as generation seeds. A complete instance that contributes to no accepted update receives $U_d=0$, while incomplete feedback contributes no observation and does not increment $n_d$.

Each accepted update event $u=(e,b,o)\in\mathcal U_e^+$ retains its reward $R_u$ and distinct contributors $\mathcal I_u$. The event identity distinguishes updates to the same object in different batches or epochs. After all batches, Eq.~\eqref{eq:practice_utility} sums an equal share $R_u/|\mathcal I_u|$ of each accepted reward over the events to which a complete instance contributed. Thus, contributions across rollouts and updates yield one utility observation $U_d$ and one increment to $n_d$ per complete practice in an epoch. This utility is a proxy for future skill improvement. Seed ranking uses the resulting statistics in Eq.~\eqref{eq:selection_score}, restricted to the eligible instances in $\mathcal T_e$.

\paragraph{Targeted generation and self-checks.}
The generation plan specifies the seed task's key challenges, data-format requirements, comparable difficulty, and aspects that can vary. It supplies these requirements without exposing private seed images, raw trajectories, or exact seed answers. Using the enabled construction, search, and image-processing tools, the generator builds or acquires the required content, derives its answer, and records supporting evidence.

Model self-checks assess the same four requirements as in Section~\ref{sec:data_evolution}: reliable answers and evidence, image-question consistency, preservation of core task challenges, and absence of answer leakage or shortcuts. Candidates that fail undergo bounded repair and renewed self-checks. Changes to the answer require renewed evidence verification. Admission also checks duplicates, label conflicts, and overlap with validation data.

\paragraph{Generation budget and next-epoch practice.}
After each nonfinal epoch, generation targets at most $\lfloor\rho M\rfloor$ admitted instances, starting with the highest-ranked eligible seeds from $\mathcal T_e$. This is an admission budget, not a prescribed fraction of generated instances in the next practice set. For each seed, generation stops at the first accepted instance or after five failed attempts. The process moves to the next unused seed and stops when the target is reached or eligible seeds are exhausted. Each successful seed contributes one accepted instance directly to the shared data bank $D$.

New instances enter the shared data bank with $A_d=n_d=0$. After generation, all statistics are discounted by $(A_d,n_d)\gets(\gamma A_d,\gamma n_d)$. At the next epoch, \textsc{Select} recomputes $\sigma_d$ over the entire expanded bank and takes the $M$ highest-ranked instances. Real and generated instances use the same rule, with $\sigma_d=+\infty$ whenever $n_d=0$. Ties are broken randomly and execution order is shuffled. Admission therefore makes a new instance available for future practice without guaranteeing selection in the next epoch. The selected practice set remains fixed throughout that epoch.

\section{Experimental Settings}
\label{app:experimental_settings}

\subsection{Tool Details}
\label{app:tool_definitions}

Table~\ref{tab:method_tools} summarizes tool capabilities and their main inputs for task solving and data generation. Concrete callable names and argument schemas are supplied by the enabled tool interfaces. Table~\ref{tab:benchmark_tools} specifies which tools are enabled for each dataset. Availability does not imply that every tool is called on every data instance.

\begin{table}[t]
\centering
\caption{Tool capabilities and main inputs used in the experiments.}
\label{tab:method_tools}
\footnotesize
\setlength{\tabcolsep}{4pt}
\renewcommand{\arraystretch}{1.12}
\begin{tabular*}{\linewidth}{@{\extracolsep{\fill}}>{\raggedright\arraybackslash}p{0.19\linewidth}>{\raggedright\arraybackslash}p{0.36\linewidth}>{\raggedright\arraybackslash}p{0.39\linewidth}@{}}
\toprule
\textbf{Tool} & \textbf{Description} & \textbf{Main inputs} \\
\midrule
\rowcolor[HTML]{E7ECF8}
\multicolumn{3}{@{}l}{\textit{Shared by execution and generation}} \\
Web Search & Search the web via Serper API for titles, URLs, and text snippets. & \texttt{query} (str, required): query. \texttt{max\_results} (int, optional): limit. \\
\midrule
Image Search & Search related images via Serper image/Lens. ImgBB provides URLs for local reverse-image inputs. & \texttt{search\_type} (str): text or reverse. \texttt{query} (str): text. \texttt{image\_url} (str): reverse. \texttt{max\_results} (int, optional). \\
\midrule
Visit & Extract the main textual content of a webpage through the Jina Reader API. & \texttt{url} (str, required): page URL. \texttt{goal} (str): information to find. \\
\midrule
Code Interpreter & Stateful Jupyter kernel for Python image processing (PIL/OpenCV), calculations, and data manipulation. & \texttt{code} (str, required): Python code. \\
\midrule
\rowcolor[HTML]{E7ECF8}
\multicolumn{3}{@{}l}{\textit{Generation only}} \\
Image Generation & Create an image from a text description with the GPT-Image-2 model. & \texttt{prompt} (str, required): image description. \\
\midrule
Image Editing & Modify or extend an existing image with the GPT-Image-2 model. & \texttt{image} (image, required): source image. \texttt{instruction} (str, required): edit. \\
\midrule
Webpage Capture & Render a public webpage in Playwright/Chromium and capture the page or a selected region. & \texttt{url} (str, required): page URL. \texttt{region} (optional): area to capture. \\
\bottomrule
\end{tabular*}
\end{table}

\subsection{Dataset Details}
\label{app:dataset_settings}

TIRBench, MMSearch-Plus, and MMBrowseComp are the main evaluation datasets. We use training data for practice, validation data to compare candidate skill-bank updates, and held-out test data to evaluate the frozen bank. VisualToolBench and AgentVista are transfer targets: we evaluate the source-trained bank on their test data without target-domain practice or generation. Table~\ref{tab:dataset_splits} reports the original dataset sizes, selected partitions, and sampling rules. Table~\ref{tab:benchmark_tools} reports tool availability. A dash marks a partition or tool unavailable in the reported experiment. The TIRBench validation pool excludes the training IDs. The MMSearch-Plus source contains one row without a reference answer. This row is excluded from all three partitions.

\begin{table}[t]
\centering
\caption{Dataset domains, sizes, and partitions. All random sampling uses seed 42.}
\label{tab:dataset_splits}
\footnotesize
\setlength{\tabcolsep}{3pt}
\renewcommand{\arraystretch}{1.15}
\begin{tabular*}{\linewidth}{@{\extracolsep{\fill}}>{\raggedright\arraybackslash}p{0.17\linewidth}>{\raggedright\arraybackslash}p{0.18\linewidth}rrrr>{\raggedright\arraybackslash}p{0.28\linewidth}@{}}
\toprule
\textbf{Dataset} & \textbf{Domain} & \textbf{Total} & \textbf{Train} & \textbf{Val.} & \textbf{Test} & \textbf{Sampling strategy} \\
\midrule
\rowcolor[HTML]{E7ECF8}
\multicolumn{7}{@{}l}{\textit{Visual Agentic Tool Use}} \\
TIRBench & Tool-Integrated Reasoning & 1,215 & 195 & 630 & 390 & Balanced random sampling across 13 task types. \\
VisualToolBench & Hybrid Tool Reasoning & 1,204 & -- & -- & 150 & Balanced random sampling of two single-turn types. \\
\midrule
\rowcolor[HTML]{E7ECF8}
\multicolumn{7}{@{}l}{\textit{Multimodal Search}} \\
MMSearch-Plus & Multimodal Search & 311 & 100 & 110 & 100 & Random Sampling \\
MMBrowseComp & Multimodal Browsing & 400 & 100 & 150 & 150 & Random Sampling \\
\midrule
\rowcolor[HTML]{E7ECF8}
\multicolumn{7}{@{}l}{\textit{Comprehensive}} \\
AgentVista & Ultra-challenging Tasks & 209 & -- & -- & 209 & All samples for testing. \\
\bottomrule
\end{tabular*}
\end{table}

\begin{table}[t]
\centering
\caption{Tool availability by dataset. Shared tools are available for execution on each marked dataset and for generation on the three main datasets. Generation-only tools are used only in V-Gym data generation. Transfer targets have no target-domain generation.}
\label{tab:benchmark_tools}
\footnotesize
\setlength{\tabcolsep}{3pt}
\renewcommand{\arraystretch}{1.15}
\begin{tabular}{@{}>{\raggedright\arraybackslash}p{0.23\linewidth}*{7}{>{\centering\arraybackslash}p{0.095\linewidth}}@{}}
\toprule
\textbf{Dataset} & \multicolumn{4}{c}{\textbf{Shared tools}} & \multicolumn{3}{c}{\textbf{Generation only}} \\
\cmidrule(lr){2-5}\cmidrule(lr){6-8}
 & Code & Web & Image & Visit & Gen. & Edit & Capture \\
\midrule
\rowcolor[HTML]{E7ECF8}
\multicolumn{8}{@{}l}{\textit{Visual Agentic Tool Use}} \\
TIRBench & $\checkmark$ & -- & -- & -- & $\checkmark$ & $\checkmark$ & -- \\
VisualToolBench & $\checkmark$ & $\checkmark$ & -- & $\checkmark$ & -- & -- & -- \\
\midrule
\rowcolor[HTML]{E7ECF8}
\multicolumn{8}{@{}l}{\textit{Multimodal Search}} \\
MMSearch-Plus & $\checkmark$ & $\checkmark$ & $\checkmark$ & $\checkmark$ & -- & -- & $\checkmark$ \\
MMBrowseComp & $\checkmark$ & $\checkmark$ & $\checkmark$ & $\checkmark$ & -- & -- & $\checkmark$ \\
\midrule
\rowcolor[HTML]{E7ECF8}
\multicolumn{8}{@{}l}{\textit{Comprehensive}} \\
AgentVista & $\checkmark$ & $\checkmark$ & -- & $\checkmark$ & -- & -- & -- \\
\bottomrule
\end{tabular}
\end{table}

\begin{table}[t]
\centering
\caption{Parameter settings for V-Gym.}
\label{tab:appendix_hyperparameters}
\footnotesize
\setlength{\tabcolsep}{4pt}
\renewcommand{\arraystretch}{1.02}
\begin{tabular*}{\linewidth}{@{\extracolsep{\fill}}>{\raggedright\arraybackslash}p{0.35\linewidth}>{\raggedright\arraybackslash}p{0.26\linewidth}>{\raggedright\arraybackslash}p{0.33\linewidth}@{}}
\toprule
\textbf{Parameter} & \textbf{Value} & \textbf{Description} \\
\midrule
\rowcolor[HTML]{E7ECF8}
\multicolumn{3}{@{}l}{\textit{Models and execution settings}} \\
Solver model & Evaluated base model & Backbone used to solve tasks. \\
Judge model & GPT-5.5 (default) & Scores complete trajectories. \\
Embedding model & text-embedding-3-small & Similarity-based retrieval. \\
Test rollouts per instance & 3 & Independent test-time solutions. \\
Solver temperature & 0.6 & Practice, comparative validation, and test sampling. \\
Solver top-$p$ & 1.0 & Nucleus sampling cutoff. \\
Maximum solver turns & 20 & Interaction turns per rollout. \\
Maximum output tokens per turn & 4,096 & Solver completion budget per turn. \\
Maximum images per instance & 100 & Image input and tool-result limit. \\
\midrule
\rowcolor[HTML]{E7ECF8}
\multicolumn{3}{@{}l}{\textit{V-Gym: skill evolution}} \\
Practice epochs $E$ & 2 & Default epochs with $M$ practice instances each. \\
Practice size $M$ & $|D|$ at initialization & Fixed number of practice instances per epoch. \\
Rollouts per instance $N$ & 2 & Trajectories for each practice instance. \\
Batch size & 5 & Practice instances per skill-update batch. \\
Specific skills per instance & At most 1 & Selected from the router document. \\
Grouping similarity threshold & 0.6 & Source-instance similarity for grouping unmatched trajectories. \\
Validation retrieval per source & 3 & Neighbors retrieved for each source instance. \\
Validation panel cap & 15 & Distinct instances per candidate update. \\
Validation solver temperature & 0.6 & Matched comparison of candidate updates. \\
\midrule
\rowcolor[HTML]{E7ECF8}
\multicolumn{3}{@{}l}{\textit{V-Gym: data evolution}} \\
Generator Model & Evaluated base model & Builds and checks candidate data instances. \\
Exploration coefficient $\beta$ & 0.1 & Exploration weight in the selection score $\sigma_d$. \\
Statistics discount $\gamma$ & 0.8 & Epoch-wise discount of instance statistics. \\
Generation budget $\rho$ & 0.6 & Admission budget relative to $M$. \\
Generation attempts per seed & At most 5 & Candidate attempts before trying another seed. \\
\bottomrule
\end{tabular*}
\end{table}

In Table~\ref{tab:benchmark_tools}, Code, Web, Image, Gen., Edit, and Capture abbreviate the corresponding tools in Table~\ref{tab:method_tools}. TIRBench generation may create new image pixels. On MMSearch-Plus and MMBrowseComp, generated images must originate from public webpages. Code Interpreter may process acquired pixels but may not draw a replacement scene.

\subsection{Parameter Settings}
\label{app:comparison_protocol}
\label{app:hyperparameters}

\paragraph{Training and validation.}
To ensure a fair comparison, we match the total training rollout budget across methods to $B_{\mathrm{train}}=EMN$, where $M$ is the initial training-set size, $E=2$, and $N=2$. V-Gym performs $E$ practice epochs with $M$ selected instances per epoch and $N$ rollouts per instance. SkillOPT performs $E$ epochs over the training set with $N$ rollouts per instance in each epoch. Ace-Skill performs $EM$ prioritized sampling draws with $N$ rollouts per draw. XSkill uses a single training pass with $EN$ rollouts per instance. Each method retains its own sampling and update mechanisms. For methods that use held-out validation, including SkillOPT and V-Gym, we evaluate each candidate update on at most 15 validation instances.

\paragraph{V-Gym settings.}
Table~\ref{tab:appendix_hyperparameters} summarizes the parameter settings for V-Gym. The solver temperature is 0.6 for practice, comparative validation, and test.

\subsection{Human Review of Evolved Data}
\label{app:human_review}

\begin{figure}[!htbp]
\centering
\includegraphics[width=\linewidth,pagebox=cropbox]{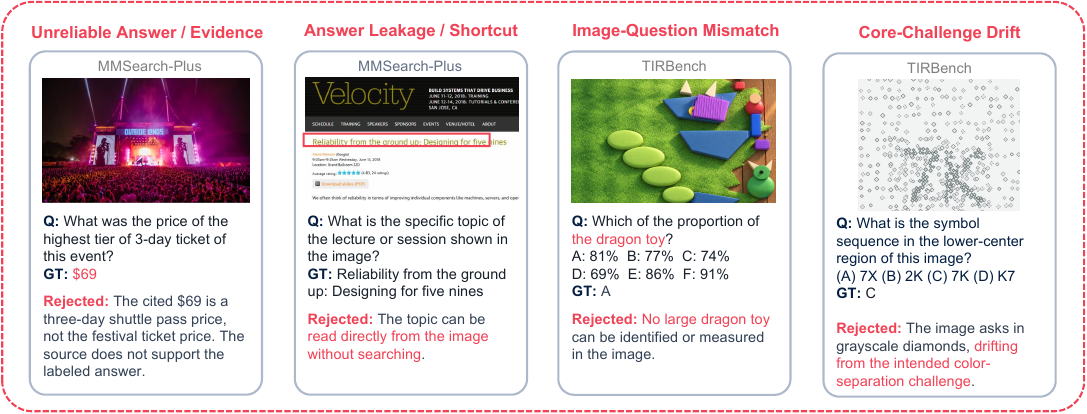}
\caption{Examples of unreliable answers or evidence, answer leakage or shortcuts, image-question mismatch, and Core-Challenge Drift, from left to right.}
\label{fig:human_review_errors}
\end{figure}

Three volunteers recruited from our institution independently reviewed the
same 100 randomly sampled, model-accepted data instances from each of TIRBench,
MMSearch-Plus, and MMBrowseComp (300 instances in total). The review assessed instance
validity, preservation of core task challenges, and comparable difficulty,
using the quality categories in Figure~\ref{fig:data_distribution}. Accepted
denotes model acceptance during self-checking, while Valid denotes instances
confirmed valid by manual review. The four error categories are
unreliable answers or evidence, image-question mismatch, Core-Challenge Drift, and answer leakage or shortcuts. Final labels were determined by
majority vote. When all three reviewers assigned
different labels, they discussed the case to reach a consensus. The mean
pairwise agreement across the three reviewer pairs was 93.56\%, measured on
the independent annotations before disagreement resolution. The final labels
were used to summarize the manual-review distributions in
Figure~\ref{fig:data_distribution}. Figure~\ref{fig:human_review_errors} provides an example of each error type.

\subsection{Examples of Valid Generated Data}
\label{app:successful_data_evolution}

\begin{figure}[!htbp]
\centering
\includegraphics[width=\linewidth,pagebox=cropbox]{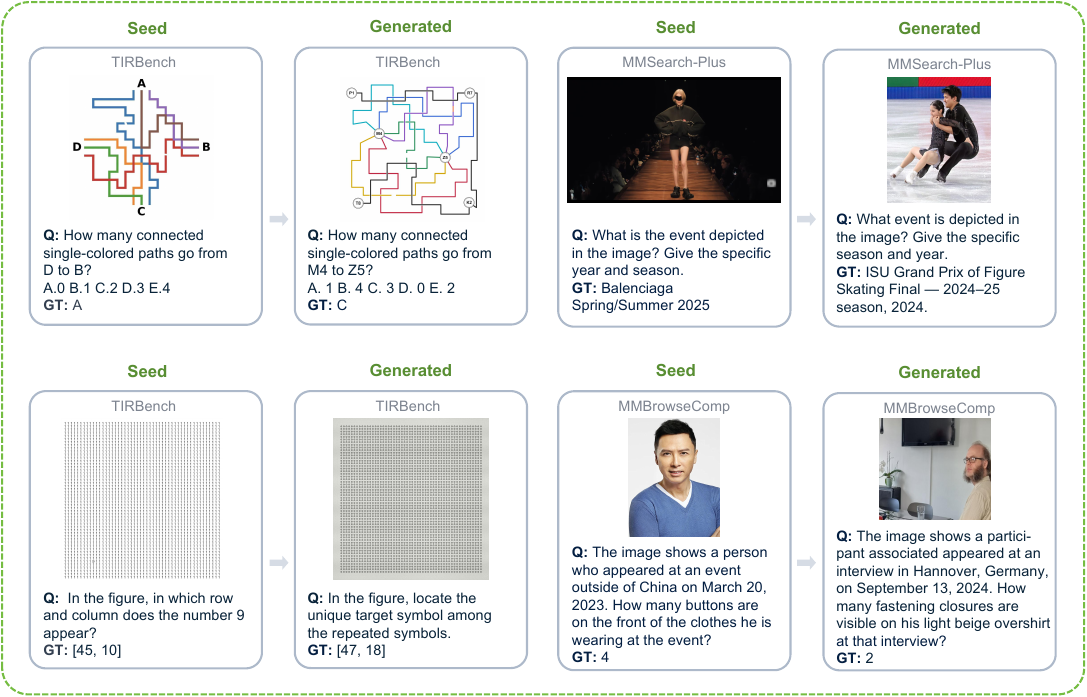}
\caption{Examples of valid generated data. Generation aims to preserve core challenges and comparable difficulty while introducing meaningful variation.}
\label{fig:successful_data_evolution}
\end{figure}

Figure~\ref{fig:successful_data_evolution} presents examples of valid generated data, illustrating the goal of preserving core reasoning challenges and comparable difficulty while introducing meaningful variation. Trajectory-derived generation plans distinguish the demands that define a task's difficulty from aspects that can vary, keeping new practice focused on the capabilities that require refinement. This combination broadens the range of practice around identified bottlenecks while retaining the challenges that make it useful. The resulting variation provides opportunities to consolidate reusable strategies and test their applicability beyond the original instances, supporting continued skill refinement.

\section{Additional Experiments}
\label{app:additional_experiments}

\subsection{Test-Time Variability}
\label{app:test_time_variability}

For each setting in Table~\ref{tab:main_benchmark_results}, we compute the sample standard deviation (SD) of single-run accuracy across three independent test runs on each benchmark, using the denominator $3-1=2$.
Mean SD is the unweighted average of these three benchmark-level SDs, reported in percentage points (pp).
The model, evaluation settings, and final skill bank, where applicable, remain fixed across test runs.
Thus, this analysis characterizes repeated evaluation with a fixed skill bank rather than variability across independent skill-evolution runs.
V-Gym has the lowest mean per-benchmark SD among the compared methods for each backbone, as shown in Table~\ref{tab:test_time_variability}.

\begin{table}[!htbp]
\centering
\caption{Mean per-benchmark sample SD of single-run accuracy (pp) for Table~\ref{tab:main_benchmark_results}. This differs from the SD of the across-benchmark mean. Dashes denote settings not evaluated.}
\label{tab:test_time_variability}
\setlength{\tabcolsep}{7pt}
\begin{tabular}{lccc}
\toprule
\multirow{2}{*}{\textbf{Method}} & \multicolumn{3}{c}{\textbf{Mean SD (pp)}} \\
\cmidrule(lr){2-4}
& \textbf{GPT-5.5} & \textbf{Gemini-3.5-Flash} & \textbf{Qwen-3.7-Flash} \\
\midrule
Baseline & 3.16 & 3.27 & 1.66 \\
Vanilla Tools & 1.29 & 2.19 & 2.32 \\
XSkill & 2.46 & --- & --- \\
Ace-Skill & 2.23 & --- & --- \\
SkillOPT & 2.09 & --- & --- \\
\rowcolor[HTML]{E7ECF8}
V-Gym (Ours) & \textbf{1.20} & \textbf{1.90} & \textbf{1.57} \\
\bottomrule
\end{tabular}
\end{table}

\subsection{Data-Evolution Hyperparameters}
\label{app:data_evolution_hyperparameters}

\begin{figure}[!htbp]
\centering
\includegraphics[width=\linewidth]{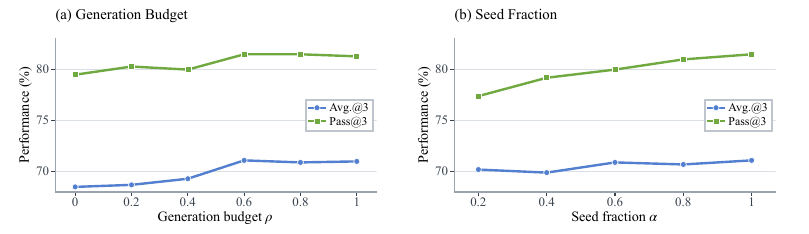}
\caption{Data-evolution analyses on TIRBench with GPT-5.5 at epoch 2. We compare (a) generation budgets with all selected seeds retained and (b) seed coverage at a fixed generation budget of $\rho=0.6$.}
\label{fig:additional_data_evolution}
\end{figure}

\paragraph{Generation budget.}
To assess how much data expansion supports skill evolution, we vary the generation budget $\rho$ while retaining all selected seeds.
The budget specifies the target number of admitted data instances relative to the $M$ practice instances per epoch.
Figure~\ref{fig:additional_data_evolution}(a) shows that both metrics improve overall as the budget increases to $\rho=0.6$, with Avg.@3/Pass@3 gains of 2.6/2.0 points over no data evolution.
Larger budgets yield similar performance, indicating diminishing returns in this range.
These results support the benefit of expanding targeted practice data and show that the default budget of $\rho=0.6$ captures the best or tied-best performance in this sweep.

\paragraph{Seed coverage.}
To test whether broader seed coverage improves the resulting skill bank, we fix the generation budget at $\rho=0.6$ and retain a random fraction $\alpha$ of the selected seeds.
The total budget remains fixed, with $1/\alpha$ new data instances per retained seed on average.
This analysis varies the number of instances generated from each seed. The default method admits one new instance per successful seed.
Figure~\ref{fig:additional_data_evolution}(b) shows that Pass@3 rises steadily with broader coverage, gaining 4.1 points from $\alpha=0.2$ to $\alpha=1$.
Avg.@3 varies within a narrower range and is also highest at full coverage.
These results support V-Gym's use of a broad set of seeds for targeted data generation, with the clearest benefit in the fraction of test instances solved across repeated attempts.

\section{Case Studies}
\label{app:skill_examples}

To illustrate the structure of our hierarchical Skill Bank, we provide examples
of the global skill, router, and a specific skill used at test time.
This Skill Bank is learned by GPT-5.5 on TIRBench.

\begin{promptbox}{Global Skill: \texttt{global\_skill.md}}
Shared tool-use and visual-evidence rules across task families.

\promptfield{Rules}
\begin{enumerate}
\setlength{\itemsep}{3pt}
\setlength{\parsep}{0pt}
\item \textbf{Verify visual evidence.} Cross-check crops, enhancements,
masks, and other transformed views against the original image or an
independently validated representation. Compare plausible alternatives.
\item \textbf{Preserve input fidelity.} Check coordinate mappings and
retain the completeness, ordering, symbols, and duplicates of structured
inputs before computation.
\end{enumerate}

\promptfield{Discipline}
\begin{enumerate}
\setlength{\itemsep}{3pt}
\setlength{\parsep}{0pt}
\item \textbf{Separate four checkpoints.} Distinguish tool execution,
candidate discovery, source confirmation, and completeness validation.
\item \textbf{Resolve contradictions.} Match each tool result to the claim
it supports. Investigate conflicting representations and rerun affected
downstream reasoning after corrections.
\end{enumerate}

\promptfield{Policy}
\textbf{Audit the final answer.} Check scope coverage, duplicates,
indexing, ordering, and output format. Resolve ambiguity through targeted
inspection or preserve uncertainty conservatively.
\end{promptbox}

\begin{promptbox}{Specific Skill Router: \texttt{router.md} (sample)}
\promptfield{\texttt{skill\_0007} --- Scaled Schematic Distance}
\textbf{Use when:} Estimate straight-line separation between marked positions
in a two-dimensional diagram with a visible linear scale.\par
\textbf{Do not use when:} The task requires route-following distance, lacks a
reliable scale, or places the targets and reference at different depths.\par
\textbf{Tool need:} Code Interpreter (optional).

\smallskip\hrule\smallskip
\promptfield{\texttt{skill\_0010} --- Grid-Maze Command Validation}
\textbf{Use when:} Validate supplied directional command sequences in a
discrete maze with identifiable endpoints, grid geometry, and traversability.\par
\textbf{Do not use when:} The task requires finding a new route, continuous
movement, or special mechanics that are not explicitly modeled.\par
\textbf{Tool need:} Code Interpreter (required).

\smallskip\hrule\smallskip
\promptfield{\texttt{skill\_0016} --- Global Jigsaw Reconstruction}
\textbf{Use when:} Recover the complete layout of equal-size, fixed-orientation
numbered panels using horizontal and vertical seams and global scene continuity.\par
\textbf{Do not use when:} The task asks for one adjacent pair or requires
rotation, mirroring, resizing, or arbitrary cropping that is not modeled.\par
\textbf{Tool need:} Code Interpreter (required).

\smallskip\centerline{\textit{\ldots\ Other skill entries omitted\ \ldots}}
\end{promptbox}

\begin{promptbox}{Specific Skill: \texttt{skill\_0016.md}}
\promptfield{Applicability}
Reconstruct a complete image grid from equally sized, fixed-orientation,
numbered panels. Exclude single-adjacency tasks and unmodeled rotation,
mirroring, resizing, or arbitrary cropping.

\promptfield{Procedure}
Use Code Interpreter with Python supplied in its \texttt{code} argument for
the image-processing and numerical steps below.
\begin{enumerate}
\setlength{\itemsep}{5pt}
\setlength{\parsep}{0pt}
\item \textbf{Extract clean panels.} Infer grid dimensions and boundaries,
then crop every panel consistently and preserve its identifier. Remove or
downweight gutters, printed labels, borders, and overlays.
\item \textbf{Measure directional compatibility.} Compare every ordered
pair of distinct panels horizontally and vertically using narrow edge strips.
Use robust color differences, gradient continuity, or feature distances.
Vary strip widths when needed.
\item \textbf{Solve the global layout.} Minimize summed horizontal and
vertical seam costs while enforcing the grid dimensions and one use of
every panel. Retain several near-best complete arrangements.
\item \textbf{Resolve competing layouts.} Rescore leading candidates under
alternative seam metrics, crops, or masks. Locate disagreements, including
swaps and row, column, or cyclic shifts, and inspect their surrounding seams.
\item \textbf{Serialize the verified layout.} After the checks below pass,
read the sequence in the requested order. Return only the complete identifier sequence in the required
syntax, normally from left to right and then from top to bottom.
\end{enumerate}

\begin{samepage}
\promptfield{Check}
\begin{enumerate}
\setlength{\itemsep}{5pt}
\setlength{\parsep}{0pt}
\item \textbf{Validate the composite.} Render the candidates and inspect
long-range boundaries, subject continuity, lighting, perspective, and plausible
outer edges. Reject close alternatives using explicit structural evidence.
Neither the lowest score nor a plausible render alone is sufficient.
\item \textbf{Audit the identifiers.} Check that every identifier occurs
exactly once and matches the verified grid.
\end{enumerate}
\end{samepage}
\end{promptbox}

\section{Prompts}
\label{app:prompts}

We present six compact prompt templates aligned with the method in
Section~\ref{sec:method}. Braced placeholders denote supplied content, while images are attached
as multimodal inputs. Tool availability follows
Table~\ref{tab:benchmark_tools}, and sampling settings follow
Table~\ref{tab:appendix_hyperparameters}.
Tool names and main inputs follow Table~\ref{tab:method_tools}. Each call uses
the registered name and argument schema supplied in the enabled tool interface.
Cropping, enhancement, and measurement are operations within Code Interpreter.

\subsection{Agent Execution and Skill Use}
\label{app:execution_prompts}

\begin{promptbox}{P1. Skill Routing and Agent Execution}
\promptfield{Routing instruction}
You are a skill router for a multimodal tool-using agent. Read the question,
images, and router document. Select at most one specific skill whose workflow
and applicability conditions match the task, or select no skill. Match the
requested operation, visual difficulty, and answer requirements. A shared
object name alone is insufficient. Prefer no match over an unsuitable skill.
Do not solve the task or use a reference answer to make the routing decision.

\promptfield{Routing input and output}
Question: \promptvar{question}. Images: \promptvar{images}.
Router document: \promptvar{routerDocument}.
Each entry gives a skill identifier and title, followed by Use when,
Do not use when, and Tool need.
Return JSON with \texttt{selected\_skill} (a listed identifier or
\texttt{null}) and \texttt{reason} (a concise applicability explanation).

\promptfield{Execution instruction}
You are a visual reasoning agent. Answer the question using the supplied images
and available tools. The global skill provides general execution guidance.
Apply the selected specific skill only where its conditions hold. Inspect the
visual evidence, choose suitable tools, and check the returned observations
before drawing conclusions. Treat a proposed action as unexecuted until a tool
result is available. Use additional inspection or verification when evidence
is ambiguous. Do not invent observations or treat skill text as evidence for
the current answer. The global skill uses Rules, Discipline, and Policy, while
the specific skill uses Applicability, Procedure, and Check. Respect these
sections, the required answer format, and the supplied interaction budget.

\promptfield{Execution input and output}
Question and images: \promptvar{question}, \promptvar{images}.
Skill context: \promptvar{globalSkill}, \promptvar{selectedSpecificSkill}.
For a null route, the specific context is empty. Retain the global skill.
Available tools: \promptvar{tools}.
For Python-based image processing and computation, call Code Interpreter
with the \texttt{code} argument. Invoke other enabled tools using their
supplied names and argument schemas. Inspect each returned result before
continuing. Return the final answer inside \texttt{<answer>...</answer>}.
Skill-specific output requirements apply to the content inside these tags.
The controller records the execution trajectory.
Do not fabricate a retrospective tool log.
\end{promptbox}

\subsection{Skill Evolution}
\label{app:skill_prompts}

\begin{promptbox}{P2. Trajectory-Grounded Skill Reflection}
\promptfield{Instruction}
Analyze the supplied trajectory group and propose reusable skill patches.
All trajectories were collected under the same batch-start skill bank.
Compare successful and failed attempts without forcing a majority conclusion.
Identify useful workflows, missing visual evidence, reasoning errors, weak
verification, and inappropriate applicability conditions. Distinguish tool
execution, the evidence it returns, and final-answer correctness. Ground every
proposed change in the supplied records and report uncertainty where evidence
is incomplete.

A patch must address a reusable execution or selection problem. Generalize
incidental entities, scenes, values, and answers into meaningful roles while
preserving constraints that define the method. Keep evidence identifiers in
provenance fields, outside reusable skill text. Do not copy exact sample
answers, benchmark labels, file paths, or one-off visual details into a skill.

\textbf{EDIT\_GLOBAL.} Revise a universal rule for evidence handling, tool use,
verification, uncertainty, or answer formatting. Do not place a task-specific
procedure in the global skill. Preserve its three sections: \textbf{Rules}
for shared evidence and input requirements, \textbf{Discipline} for tool-use
and verification conduct, and \textbf{Policy} for final-answer decisions.

\textbf{EDIT\_SPECIFIC.} Revise an existing skill's procedure, checks, or applicability
conditions. Specify the exact target skill, operation, text location, and
revised content. State positive and negative applicability boundaries when
changing its scope.

\textbf{CREATE.} Propose a distinct reusable skill when the bank lacks the
required workflow. Provide a complete procedure and applicability metadata,
rather than a description of this individual solution.

Organize each specific skill into \textbf{Applicability}, \textbf{Procedure},
and \textbf{Check}. Applicability states positive and negative scope boundaries.
Procedure includes the core strategy, tool workflow, required evidence,
intermediate outputs, and final-answer policy. Check specifies verification
requirements, common failure modes, risk controls, and fallback conditions
where applicable. Metadata must include a concise title,
applicability summary, positive and negative triggers, tool requirements,
answer format, and retrieval description.
Express tool steps using the names and main inputs in
Table~\ref{tab:method_tools}. Image-processing steps use Code Interpreter
with Python in its \texttt{code} argument.
The controller renders the approved metadata into router entries, each headed
by the skill identifier and title and containing \textbf{Use when},
\textbf{Do not use when}, and \textbf{Tool need}.
Tool need names the relevant tools from Table~\ref{tab:method_tools} and states
whether they are required or optional.

Return no patch when the evidence does not justify a reusable change. Do not
predict validation gains or decide whether an update should be accepted.

\promptfield{Input}
Batch-start global skill: \promptvar{globalSkill}.
Specific skills and metadata: \promptvar{specificSkills}.
Group questions, images, trajectories, route decisions, and evaluation
feedback: \promptvar{groupEvidence}.
Permitted evidence identifiers: \promptvar{evidenceIds}.

\promptfield{Output}
Return JSON with \texttt{observations} (evidence-grounded findings) and
\texttt{patches} (a list, possibly empty).
Each patch contains \texttt{branch}, \texttt{target\_object},
\texttt{operation}, \texttt{location}, \texttt{content},
\texttt{metadata\_changes}, \texttt{evidence\_refs}, and \texttt{rationale}.
For \texttt{CREATE}, the content and metadata fields contain the complete
skill and its metadata. Use a local proposal identifier for its target.
Use \texttt{append}, \texttt{insert\_after}, or \texttt{replace} for edits,
and \texttt{create} for a new skill.
\end{promptbox}

\begin{promptbox}{P3. Object-Level Patch Consolidation}
\promptfield{Instruction}
Consolidate the supplied patches targeting the same skill object into one
candidate update. Patches may originate from different trajectory groups,
but refer to the same batch-start bank. Use the supplied batch-start object
as the common base. Do not incorporate other candidate updates.

Deduplicate equivalent edits, integrate complementary changes, and resolve
conflicts using their supporting evidence. Preserve useful operational
constraints, verification rules, and applicability boundaries. When evidence
does not resolve a conflict, retain the supported existing rule and identify
the unresolved issue. Do not invent additional changes.

For the global skill, preserve the Rules, Discipline, and Policy sections.
For a specific skill, consolidate edits to Applicability, Procedure, and Check
together, preserving this organization. Its resulting metadata must describe
the revised skill consistently and retain the router fields specified in P2.
For a proposed new skill, check that its procedure and metadata are complete
and mutually consistent. Keep tool names and inputs consistent with
Table~\ref{tab:method_tools}.

Keep provenance separate from reusable skill text. Retain references to the
patches and evidence supporting the consolidated change. The controller
verifies these references and computes the union of distinct contributing
instances. Repeated trajectories from one instance do not create additional
contributors.

\promptfield{Input}
Target object: \promptvar{targetObject}.
Batch-start content and metadata: \promptvar{batchStartObject}.
Cross-group patches with provenance: \promptvar{objectPatches}.
For a new object, the batch-start content is empty.

\promptfield{Output}
Return one JSON record with \texttt{target\_object},
\texttt{updated\_content}, \texttt{updated\_metadata},
\texttt{source\_patch\_refs}, \texttt{evidence\_refs}, and
\texttt{unresolved\_conflicts}.
This record proposes a candidate for comparative validation. It does not
decide acceptance.
\end{promptbox}

\subsection{Data Evolution}
\label{app:data_prompts}

\begin{promptbox}{P4. Trajectory-Guided Generation Planning}
\promptfield{Instruction}
Analyze the seed task and its execution trajectories to prepare a reusable
generation plan. Identify the seed question, its answer format, and the visual
or external evidence needed to solve it. Compare successful and failed
attempts to locate consequential differences in perception, tool use,
reasoning, and verification. Distinguish observed successful operations from
hypotheses inferred from failed attempts. Do not present an unexecuted
procedure as demonstrated evidence.

Specify the task's core challenges and explain how they affect the solution
process. Target comparable difficulty by preserving these challenges and
their relevant evidence relationships. Describe feasible construction or
acquisition directions, answer-consistency requirements, and checks of the
final artifact. Do not equate visual clutter, object count, image size, or
additional tool calls with difficulty without a task-specific justification.

Preserve the core challenges and the evidence relationships essential to them.
Task types, answer formats, objects, scenes, layouts, wording, and answer values
may vary when they are not essential to these challenges. For each proposed
generation direction, specify a coherent question, answer format, and evidence
requirements.

Produce an abstract public plan for the generator. Replace concrete seed
identities and answer-bearing details with semantic roles and placeholders.
Do not include private seed images, raw trajectories, exact seed answers,
local paths, or copied seed-specific text. The plan must be feasible with the
enabled tools and support generation without access to the private seed.

\promptfield{Input}
\promptvar{seedTask}, \promptvar{seedImages},
\promptvar{executionTrajectories}, \promptvar{evaluationFeedback},
and \promptvar{enabledTools}.

\promptfield{Output}
Return JSON with \texttt{private\_analysis} (observed evidence, bottlenecks,
and difficulty rationale) and \texttt{public\_plan}.
The public plan contains \texttt{task\_contract}, \texttt{core\_challenges},
\texttt{difficulty\_rationale}, \texttt{generation\_strategy},
\texttt{allowed\_variations}, and \texttt{verification\_requirements}.
The \texttt{task\_contract} defines the question, answer format, and evidence
requirements for the proposed generation direction.
Only the public plan is passed to the generator. Retain the private analysis
for self-checking.
\end{promptbox}

\begin{promptbox}{P5. Targeted Data Generation and Repair}
\promptfield{Instruction}
Create one candidate practice instance from the public generation plan.
Preserve its core challenges, essential evidence relationships, and
comparable-difficulty target. Task type and answer format may differ from the
seed when they are not essential to these challenges. Follow the question,
answer format, and evidence requirements specified for the chosen generation
direction. Treat proposed scenes, viewpoints, and layouts as adaptable unless
they are essential to the core challenges.

Use only enabled tools from Table~\ref{tab:method_tools}: Image Generation,
Image Editing, Webpage Capture, and the shared execution tools.
Follow the supplied name and argument schema for every call. Use Code
Interpreter with the \texttt{code} argument for image processing and calculations. When
construction is enabled, create new answer-bearing content and apply suitable
transformations. When web acquisition is required, find and capture actual
source content, preserve its provenance, and process it only as permitted.
Do not claim that a source supports the candidate merely because it appeared
in the tool history. Do not request or reuse private seed pixels or answers.

Inspect the artifacts as construction proceeds. Derive the answer from the
final artifact and supporting evidence, rather than assuming that an intended
edit or earlier intermediate state determines it. For counts, regions,
coordinates, measurements, or other computed answers, recompute the result
after transformations that may affect it. Record enough evidence to verify
the submitted question and answer.

Check for ambiguity, unintended cues, missing evidence, and changes that
remove the core challenge. Revise or restore a usable intermediate artifact
when needed. Keep the question consistent with the visible content and require
the intended solution process.

If repair feedback is supplied, continue the same candidate with its existing
artifacts, verified sources, and generation context. Address the defect with
the smallest effective change. Replace content when necessary to obtain a
valid instance. Do not simplify away the prescribed challenge to pass a
check. Recheck affected evidence after each repair. Any answer change requires
renewed verification before resubmission.

\promptfield{Input}
\promptvar{publicPlan}, \promptvar{enabledTools},
\promptvar{existingArtifacts} (empty for initial generation),
and optional \promptvar{repairFeedback}.

\promptfield{Output}
Use the enabled tool interfaces to produce and submit a candidate with
\texttt{images}, \texttt{question}, \texttt{answer},
\texttt{supporting\_evidence}, and \texttt{source\_provenance}.
Persist the referenced artifacts. Submission starts self-checking. It does
not itself admit the candidate to the data bank.
\end{promptbox}

\begin{promptbox}{P6. Generated-Data Self-Check}
\promptfield{Instruction}
Assess the actual candidate images, question, answer, and supporting evidence.
Use the original seed analysis as the reference for core challenges and difficulty.
Construction plans guide generation. Their proposed surface details and claims
are not proof that the final candidate is valid. Check four requirements,
giving a concise evidence-based judgment for each.

\textbf{1. Reliable answers and evidence.}
Verify that the answer follows from the final artifact and identifiable
supporting evidence. Recheck calculations, counts, spatial assignments, and
source relationships when relevant. Distinguish verified facts from
assumptions. Identify unsupported or ambiguous answer components and evidence
that refers only to an earlier version.

\textbf{2. Image--question consistency.}
Confirm that the question refers to content present in the images, its
instructions and answer format are well defined, and the answer addresses
it. Check whether transformations or repairs changed the visual evidence
without a corresponding update to the question or answer.

\textbf{3. Preservation of core task challenges.}
Determine whether solving the candidate still exercises the intended
challenges. Justify comparable difficulty through the required solution
process, bottlenecks, and evidence relationships. State uncertainty when the
comparison is unsupported. Appearance, metadata, a planned tier, or a fixed
number of objects or tool calls does not establish difficulty.

\textbf{4. Absence of answer leakage or shortcuts.}
Check for exposed answers, unintended annotations, copied answer-bearing seed
content, and cues that bypass the intended challenge.

Accept only when all four requirements are supported, including the
comparable-difficulty target. For a repairable failure or missing evidence,
identify what must be verified or revised and what usable content can remain.
Preserve the candidate's task contract and core challenges. Request renewed
verification whenever the answer or its evidence changes. Keep repair feedback public: do not disclose
private seed identities, exact answers, or copied seed content. Reject
unrepairable candidates. The controller also rejects unresolved candidates
when the repair budget is exhausted.

\promptfield{Input}
\promptvar{candidate} and \promptvar{supportingEvidence}.
Private reference: \promptvar{seedTask}, \promptvar{seedImages}, and
\promptvar{privateAnalysis}.

\promptfield{Output}
Return JSON with \texttt{decision} (\texttt{accept}, \texttt{revise}, or
\texttt{reject}), four \texttt{check\_judgments} (each with a verdict and
evidence), \texttt{difficulty\_comparison}, and \texttt{repair\_feedback}.
Use empty feedback on acceptance. Otherwise, state the defect and the required
action, or why repair is infeasible.
\end{promptbox}

\end{document}